%% file: paper.tex
\documentclass[runningheads]{llncs}
\usepackage[utf8]{inputenc}
\usepackage[]{hyperref}

\usepackage{graphicx}
\usepackage{booktabs}
\usepackage{url}
\usepackage[%
  square,        
  comma,         
  numbers,       
  sort&compress, 
]{natbib}
\usepackage{subcaption}
\usepackage{caption}
\usepackage{xcolor}
\usepackage{tikz,float}
\usepackage{xparse}

\usepackage{amsmath,amssymb,amsthm,mathrsfs,amsfonts,dsfont} 

\usepackage{algorithm}
\usepackage{algorithmic}

\usepackage[activate={true,nocompatibility},final,tracking=true,kerning=true,spacing=true,factor=1100,stretch=10,shrink=10]{microtype}

\makeatletter
\newcommand*{\centerfloat}{%
  \parindent \z@
  \leftskip \z@ \@plus 1fil \@minus \columnwidth
  \rightskip\leftskip
  \parfillskip \z@skip}
\makeatother

\definecolor{paper_blue}{RGB}{85,114,175}
\definecolor{paper_orange}{RGB}{209,139,85}

\NewDocumentCommand{\statcirc}{ O{#2} m }{%
    \begin{tikzpicture}
    \fill[#2] (0,0) circle (0.7ex); 
    \end{tikzpicture}
  }

\theoremstyle{definition}

\newcommand\independent{\protect\mathpalette{\protect\independenT}{\perp}}
\def\independenT#1#2{\mathrel{\rlap{$#1#2$}\mkern2mu{#1#2}}}

\usepackage[draft,inline,marginclue,index]{fixme}

\fxsetup{theme=color,mode=multiuser}
\FXRegisterAuthor{bettina}{abettina}{\color{green}Bettina}
\FXRegisterAuthor{pieter}{apieter}{\color{violet}Pieter}
\FXRegisterAuthor{paul}{apaul}{\color{orange}Paul}
\FXRegisterAuthor{gilles}{agilles}{\color{blue}Gilles}

\date{}

\begin{document}

\title{Ethical Adversaries: Towards Mitigating Unfairness with Adversarial Machine Learning}
\titlerunning{Ethical Adversaries}
\authorrunning{Pieter Delobelle et al.}


\author{Pieter Delobelle\inst{1} \and Paul Temple\inst{2} \and Gilles Perrouin\inst{2} \and\\ Benoît Frénay\inst{2} \and Patrick Heymans\inst{2} \and Bettina Berendt\inst{1,3}}

\institute{Department of Computer Science, KU Leuven and Leuven.ai,\\
\texttt{firstname.lastname@kuleuven.be}
\and
PReCISE, NaDi, Université de Namur,\\
\texttt{firstname.lastname@unamur.be}
\and
Faculty of Electrical Engineering and Computer Science, TU Berlin}

\maketitle              

\begin{abstract}
\input{absrtact}

\end{abstract}

\section{Introduction}\label{sec:intro}
Machine learning eases the deployment of systems that tackles various tasks: spam filtering, image recognition, etc.
One of the most trendy applications is decision support. 
These systems give recommendations on who should get a loan, predict who could commit subsequent offences, etc. based on data describing the individuals affected.
Such systems have a desirable property: they provide objective, supposedly consistent decisions based on a collection of data.
At first glance, this could counteract unfair decisions made by humans.

However, these support systems still exhibit unfair behaviour.
Such behaviours can possibly impact certain individuals, belonging to social or even protected groups.
Well-studied examples include the COMPAS system that predicts the recidivism of pre-trial inmates~\citep{angwinMachineBiasThere2016, chouldechovaFairPredictionDisparate2017} and keep taking decisions in favor of Caucasian people compared with African-Americans.
We consider fairness where the impact on individuals can be categorized as either \emph{allocational harm} or \emph{representational harm}~\citep{blodgettLanguage2020}.
With allocational harm, the favorable outcome (e.g. bail being granted) differs between social groups.
Representational harm is more subtle, and include \emph{differences in performance} between social groups, and \emph{stereotyping}.
We focus on allocational harm in this work, as decision support systems with different outcomes can affect social groups far beyond the outcome itself. 

If an allocational harm exists when advising for a favorable outcome, the decision could, in turn, affect the social group(s) who did not receive that outcome and,
ultimately, risking the creation of a feedback loop where unfair behaviour is amplified~\citep{overdorfQuestioningAssumptionsFairness2018, ensign2017}.
For example, consider a system that imposes more expensive loans to African-American people, who then fail to repay them, that will lead them to ask for another loan, etc.

Because of these consequences, researchers increasingly focus on incorporating fairness objectives in their systems.
In \emph{discrimination-aware data mining} (DADM), modifications were developed and applied to data, learning algorithms, or resulting patterns and models~\citep{DBLP:journals/tkde/HajianD13}.
More recently, adversarial fairness is continuing in this field with, for instance, research on learning representations~\citep{zemel2013learning, madrasLearningAdversariallyFair2018} and task-specific fair models~\citep{raff2018, adel2019, sattigeriFairnessGANGenerating2019}.

Adversaries are also used when assessing the security of machine learning based systems.
\citet{biggio2018} synthesised a decade of research in adversarial machine learning.
This domain of research aims at finding or creating examples that are problematic for a machine learning model, \textit{e.g.} \citet{papernot2016,papernot2017,biggio2013evasion}.
These examples can be injected directly into the training phase in order to perturb the training of the model, known as \emph{poisoning attacks}, or they can simply be used to bypass the model that is supposed to act as a filter, in this case, they are called \emph{evasion attacks}.

In this paper, we propose a new framework implementing a gray-box fairness scenario coupling evasion attacks and fair machine learning using gradient reversal.
We evaluate our framework on three datasets: (i) COMPAS, (ii) German Credit, and (iii) Adult.
We show demographic parity and equal opportunity improved when comparing to the state of the art while globally improving the model’s utility. Our framework thus reconciles fairness and model performance.

This paper is organised as follows: \autoref{sec:background} discusses 
related work on adversarial machine learning techniques but also 
on measuring and mitigating unfairness.
\autoref{sec:our-work} presents 
our new framework, followed by its evaluation on the COMPAS, German Credit and Adult datasets in \autoref{sec:evaluation}. \autoref{sec:conclusion} concludes and gives an outlook on future work.

\section{Background and Related Work}\label{sec:background}
\subsection{Poisoning and Evasion Attacks}\label{sec:adv-ml}
Adversarial machine learning assesses the required effort to make a classifier unusable by forcing it to perform so many errors that users will not trust its predictions anymore~\citep{biggio2018}. The generation of adversarial attacks follows this black-box process: (i) probe an existing target model to gain information about it, (ii) copy an existing example, (iii) apply an adversarial technique that will modify the example depending on the desired goal. 

Various models can be attacked including support vector machines (SVMs), linear models and even neural networks (NNs)~\citep{papernot2016,papernot2017,biggio2013evasion}. Since all machine learning models are based on a similar set of assumptions, including the fact that they statistically approximate data distributions, adversarial machine learning leverage on these assumptions to train a surrogate classifier to start the attack on and results are transferred to the target model~\citep{demontis2019transfer}.
Only one restriction remains on the surrogate classifier, attacks are gradient-based techniques requiring the discriminant function to be differentiable. We distinguish between \emph{poisoning attacks} and \emph{evasion attacks}. 
In the former, malicious examples are introduced in the training set in order to significantly and permanently affect the model to be trained~\citep{biggio2012poisoning,biggio2013poisoning}.

In concurrent work by \citet{solans2020poisoning}, poisoning attacks have been used to influence the fairness of machine learning models in a black-box manner.
The authors have also linked their poisoning attack to demographic parity, an evaluation metric that will be introduced in \autoref{ss:fairness-metrics}.

\citet{kulynychPOTs2020a} also used poisoning attacks, specifically for countering effects of credit scoring systems. 
In addition, they provide an outline of how users can affect optimization systems to mitigate negative externalities, called Personal Optimization Technologies (POTs). 
This framework could also be used to ground the adversarial attacks generated by the \emph{Feeder} from our framework.

In this paper, we consider evasion attacks that are performed on \textit{an already trained model}. We craft adversarial examples that are supposed to belong to a class while the model will assign them with a different one because of specific characteristics, highlighting an unfair behavior regarding a certain population. By carefully reintroducing these examples during retraining, we hypothesize that the retrained model will be fairer. While we rely on a similar example generation technique, we have a distinct exploitation goal.

\subsection{Evaluating Fairness}\label{ss:fairness-metrics}
There exist several measures of fairness in the literature, \emph{e.g.} demographic parity~\citep{dworkFairnessAwareness2012}, equalized odds and equalized opportunity~\citep{hardtEqualityOpportunitySupervised2016}, statistical parity~\citep{feldman2015certifying, zemel2013learning}, disparate impact~\citep{chouldechovaFairPredictionDisparate2017,feldman2015certifying}, and threshold testing~\citep{pmlr-v84-pierson18a}. 
In the following, we focus on the most popular and representative measures: (i) demographic parity and (ii) equalized opportunity.
We define all measures via the predicted values of the classifier 
$\hat{Y}$ and the protected attribute $A$.
We identify the disadvantaged group with $A=1$ and the privileged group with $A=0$.
The similarities of predictions are described for $\hat{Y}=1$.

Since the focus of most fairness measures is on the disadvantaged group having fewer (desired) opportunities, $\hat{Y}=1$ is generally the desired outcome.  
One set of measures expresses the requirement that the predicted values of the classifier $\hat{Y}$ 
conditioned on the protected attribute
be equal~\citep{calders2010} or the difference to be within an acceptable range.
\begin{definition}{Demographic parity (DP).}
DP is the equality or similarity of prediction outcomes as an absolute difference~\citep{dworkFairnessAwareness2012, raff2018}:

\begin{equation}
  DP = \left|{P(\hat{Y} = 1 \mid A = 0)} - {P(\hat{Y} = 1 \mid A = 1)}\right| \leq \epsilon.
\end{equation}
\end{definition}

\begin{definition}{Demographic parity ratio (DPR).}
DPR is the equality or similarity of prediction outcomes as a ratio:

\begin{equation}
 DPR = \frac{P(\hat{Y} = 1 \mid A = 1)}{P(\hat{Y} = 1 \mid A = 0)} \geq \tau.
\end{equation}
\end{definition}
Requiring $DP=0$ or $DPR=1$ would require exact equality in the outcome predictions for both groups. This is unrealistic for most data, 
such that real-world usage of such measures is less restrictive. For instance, in a legal setting, the US Equal Employment Opportunity Commission (EEOC) uses the DP ratio with $\tau = 0.8$ (``\emph{80\% rule}'' \citep{feldman2015certifying}), stating that disparate impact caused by employment-related decisions or structures can only be ascertained if $DPR \leq 0.8$.

Demographic parity has received some criticisms, since the measure does not necessarily report on what many would define as fairness~\citep{dworkFairnessAwareness2012}. 
This issue stems from ignoring both the true outcome and individual merits. For instance, consider a selection procedure where we consider two individuals belonging to the protected group. 
Let say that one individual is qualified (\emph{i.e.}, with high chances to get a positive true outcome $Y=1$) and the other one is not.
Not selecting the qualified individual could be considered unfair, but the selection would satisfy demographic parity even when selecting the not-qualified individual.
So these \emph{token} individuals are not guaranteeing fairness since qualified individuals from the protected group are still mistreated.

Addressing the criticisms of demographic parity, \citet{hardtEqualityOpportunitySupervised2016} presented two other metrics that extend the aforementioned ones. By including the true outcome $Y$, the authors show that this variable can serve as a \emph{justification} for the predicted outcome. For example, in the case of COMPAS, this is the recidivism rate as measured by violent crimes in a two-year window. Conditioning by the true outcome is a justification that the authors consider to be a suitable interpretation of the \emph{task-specific similarity measure} from \citet{dworkFairnessAwareness2012}, which can otherwise be difficult to come up with. 
This is also very similar to \emph{disparate mistreatment}~\cite{zafarFairnessDisparateTreatment2017, barocas-hardt-narayanan} used as an evaluation metric by \citet{adel2019}.

\begin{definition}{Equal opportunity (EO).}
  EO requires an independence $\hat{Y}\independent A \mid Y$ of $\hat{Y}$ and $A$ conditioned on the true outcome $Y$. 
  Expressed as a difference, this yields:


\begin{equation}
  \left| P( \hat{Y} = 1 \mid A = 0, Y = 1) - P( \hat{Y} = 1 \mid A = 1, Y = 1) \right| \leq \nu .
\end{equation}

\end{definition}
``Equality of opportunity'' is satisfied if $\nu = 0$, and larger absolute values are indicative of 
unfairness
in the model or data. 

\subsection{Fair Neural Networks} \label{sec:fair-NN}
Fair models have been studied for a variety of learning algorithms, such as Naive Bayes classifiers~\citep{calders2010} or SVMs~\citep{zafarParityPreferencebasedNotions2017}.
Nowadays, the focus is also on neural networks due to their prediction performance~\citep{adel2019, raff2018fair, madrasLearningAdversariallyFair2018}.

Several work have tried to mitigate unfairness in neural networks with white-box adversaries~\citep{zhangMitigatingUnwantedBiases2018, edwards2015, madrasLearningAdversariallyFair2018, adel2019, raff2018fair}.
In all these instances, a new model architecture is proposed with two goals:
(i) predicting the main attribute $Y$ (which we will refer to as the \emph{utility of the model}; with $Y=1$ being the positive outcome);
(ii) not being able to predict the protected attribute $A$ (with $A=1$ considered as belonging to the protected group).
The joint goals can be formally defined as a min max optimization problem~\citep{edwards2015} over the loss function $L$, i.e., $\min_\theta \max_\phi L\left(\theta, \phi\right),$ with an adversary $\phi$ and an encoder with parameters $\theta$.
We use this representation to predict both $Y$ and $A$ via a white-box adversary and a neural network.
\citet{ganin2016domain, raff2018, adel2019} all proposed to optimize a variant of the following loss function following

\begin{equation}
  \label{eq:loss}
  L(\theta, \phi) = E_{\theta, \phi}(X, Y) - \lambda D_{\theta, \phi}(X, A),
\end{equation}
with $D_{\theta, \phi}$ the loss for predicting $A$ from $X$, and $E_{\theta, \phi}$ the loss for the target prediction $Y$ also from $X$ and $\lambda$ a hyper-parameter.

Gradient reversal was introduced by \citet{ganin2016domain} for domain adaptation, 
and later adapted by \citet{raff2018} and \citet{adel2019} who treated the protected attribute $A$ as a domain label. 
The gradient reversal strategy assumes that multiplying by a negative sign will increase the loss $D_{\theta, \phi}(X, A)$ of the branch $h_a: X \rightarrow \hat{A}$ and yields a representation $X^*$ that is maximally invariant to changes in $A$~\cite{adel2019, raff2018}. 

Using gradient reversal for fairness is based on the intuition that the inability to predict $A$ is a suitable fairness goal.
This differs slightly from the fairness evaluations presented in \autoref{ss:fairness-metrics}, but a similar loss function from \autoref{eq:loss} based on demographic parity led to the architecture of FAD~\citep{adel2019}, which leverages gradient reversal specifically for fairness.

However, there is no guarantee that gradient descent with flipped gradients does guarantee the maximal invariance required for fairness.
In the worst case, maximizing the loss $D_{\theta,\phi} (X, A)$ can even result in the opposite optimum for the shared layers with regard to $A$, because flipping the gradients with regard to $A$ makes it perform gradient ascent for $A$.
With the shared layers performing gradient ascent w.r.t. $A$ followed by gradient descent in the adversarial branch, this creates a discrepancy between the parameters defining both components for predicting $A$.
This means that the model is not only not maximally invariant on the last shared layer, but that the shared layers are still explicitly learning to predict the protected attribute $A$.

This is one of the major limitations of using GRL for fair models, as predictions of main attribute $Y$ are not made on `fair' representations.
\citet{elazar-goldberg-2018-adversarial} made an empirical observation on \emph{leakage} of protected attributes specifically for text-based classifiers that can also be traced back to this.
In \autoref{sec:our-work} we clarify how our ethical adversaries framework mitigates this issue, thus allowing GRL to be used for training fair models.

\section{Ethical Adversaries Framework}\label{sec:our-work}
\input{architecture}

\section{Evaluation}\label{sec:evaluation}
We evaluate our model on three popular datasets: COMPAS~\citep{angwinMachineBiasThere2016}, German Credit, and the Adult Census~\citep{kohavi1996scaling}.
The COMPAS dataset was originally a sample of outcomes from the COMPAS system that predicted the risk of recidivism.
This caused a debate about whether or not this score was disadvantaging African Americans~\citep{angwinMachineBiasThere2016, dieterichCOMPASRiskScales2016, chouldechovaFairPredictionDisparate2017, corbett-davies2018}. The dataset, therefore, includes the race of individuals. 
In line with other research~\citep{adel2019, angwinMachineBiasThere2016, zafarParityPreferencebasedNotions2017}, we will only use individuals from \emph{Caucasian} or \emph{African-American} descent. 
As other groups are clearly less represented (e.g., only 31 instances for people of Asian descent), this poses issues during training and evaluation. 
It implies that there are minorities that are excluded from many studies; more datasets would be needed to study whether patterns of unfairness are similar and mitigation measures can be transferred, or whether these affect different demographics differently.
COMPAS is composed of 5,278 instances and represented by 12 features.
The target variable is whether a person has recidivated within two years. The race is used as a protected attribute.

The Adult dataset gathers 32,000 instances represented by 9 features. We use gender as a protected attribute and the binary target variable is income, whether someone earns more than 50,000 USD. 
German Credit is the smallest dataset, with only 1,000 instances and 20 features. There is a class imbalance, with 70\% of all samples good credits and only 30\% bad credits. The protected attribute is age, with a threshold at 25 years.

For reproducibility purposes, we have publicly released our code and provided users with a template that they can incorporate in their projects.
It is compatible with all PyTorch models with only minor modifications, i.e., adding an adversarial branch and replacing the training loop.
We recall that we have used the secML package\footnote{https://secml.gitlab.io/} (v0.11) for running evasion attacks.

\subsection{Training setup}\label{ss:setup}

\emph{The model under attack.}
We start from a neural network of 3 hidden layers with 32 hidden units for COMPAS and German Credit and 128 for Adult, due to its larger encoded input. 
Each of the hidden units has a ReLU activation. 
This activation function is computationally efficient and mitigates the issue of vanishing gradients since the function never saturates, which makes it one of the most popular activation functions. 
For the output units, a softmax activation was used to get the classification and a linear activation for COMPAS.
The network, including the adversarial reader, is trained with the Adam optimizer with $\beta_1=0.9, \beta_2=0.9999$ and an initial learning rate $l_r = 0.01$, which is adjusted by a factor of~$0.1$ when reaching a plateau.

\emph{The adversarial reader.}
The adversarial reader is part of the model under attack and therefore follows the same training regime. 
The joint loss follows \autoref{eq:loss} by including the GRL. The individual losses for both $h_A$ and $h_y$ are binary cross-entropy loss, except for COMPAS. In that case, the risk score is predicted as a regression problem with the MSE loss and then thresholded at 4 (low \emph{vs} medium and high risk).

\emph{The adversarial feeder.}
In our setting, we can use the same training set for both the feeder and reader since they are part of the same, unique architecture.

We also approximate---relying on the earlier discussed transferability of attacks~\citep{demontis2019transfer}---the attacked model by an SVM with a radial basis function kernel. We set the hyperparameters $C$ and $\gamma$ with a grid search with a reduced number of values: respectively $\{0.0001; 0.001; 0.01; 0.1; 1.0\}$ and $\{0.01; 0.1; 1; 10; 100;1000\}$.
We performed 10-fold cross-validation.

\subsection{Mitigating unfair representations}
\begin{figure}[tbh]
  \centering
  \centerfloat
  \begin{subfigure}[t]{0.3\columnwidth}
    \includegraphics[width=\columnwidth]{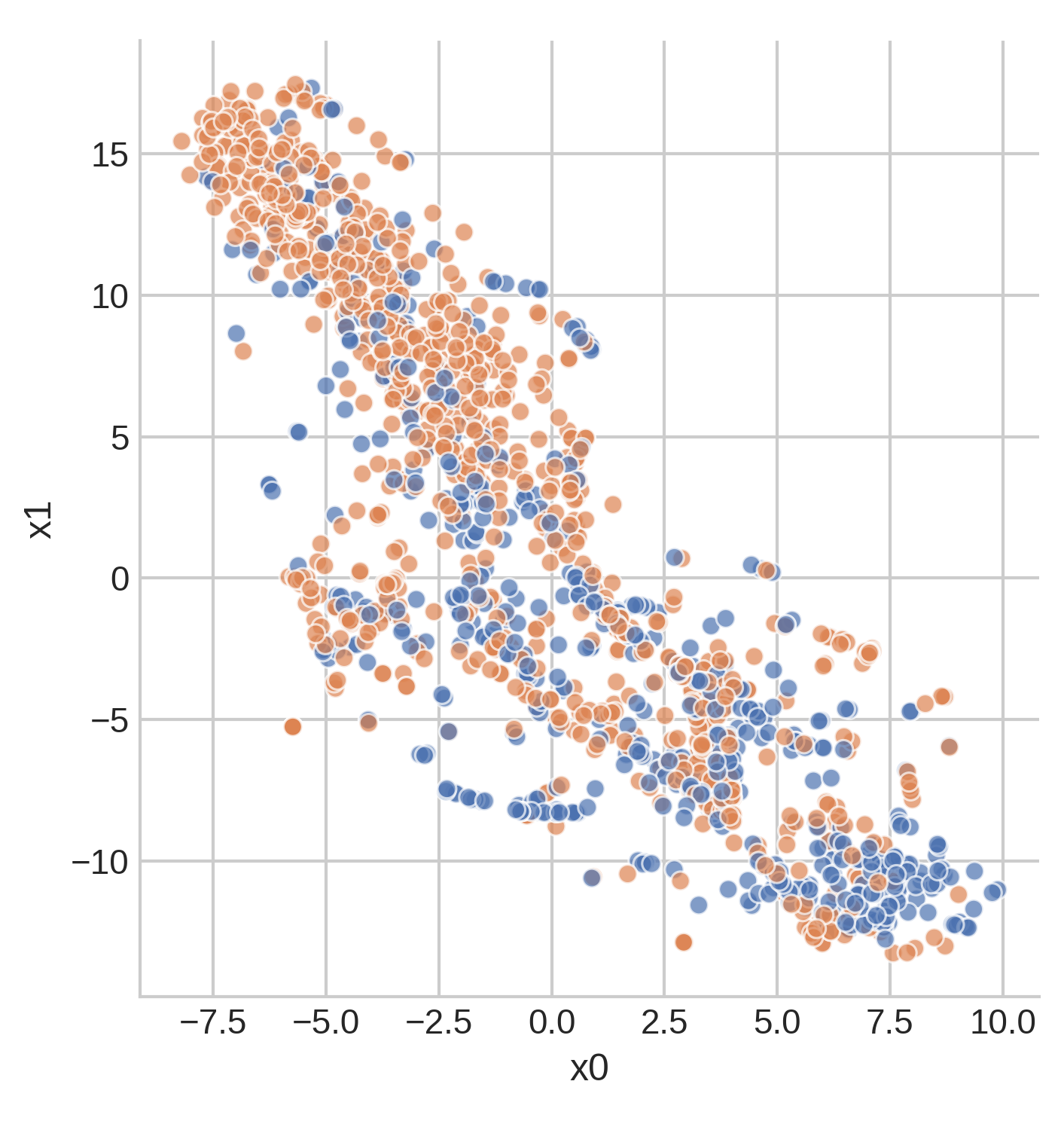}
    \caption{Naive model}
    \label{fig:tsne-biased}
  \end{subfigure}
  ~
  \begin{subfigure}[t]{0.3\columnwidth}
    \includegraphics[width=\columnwidth]{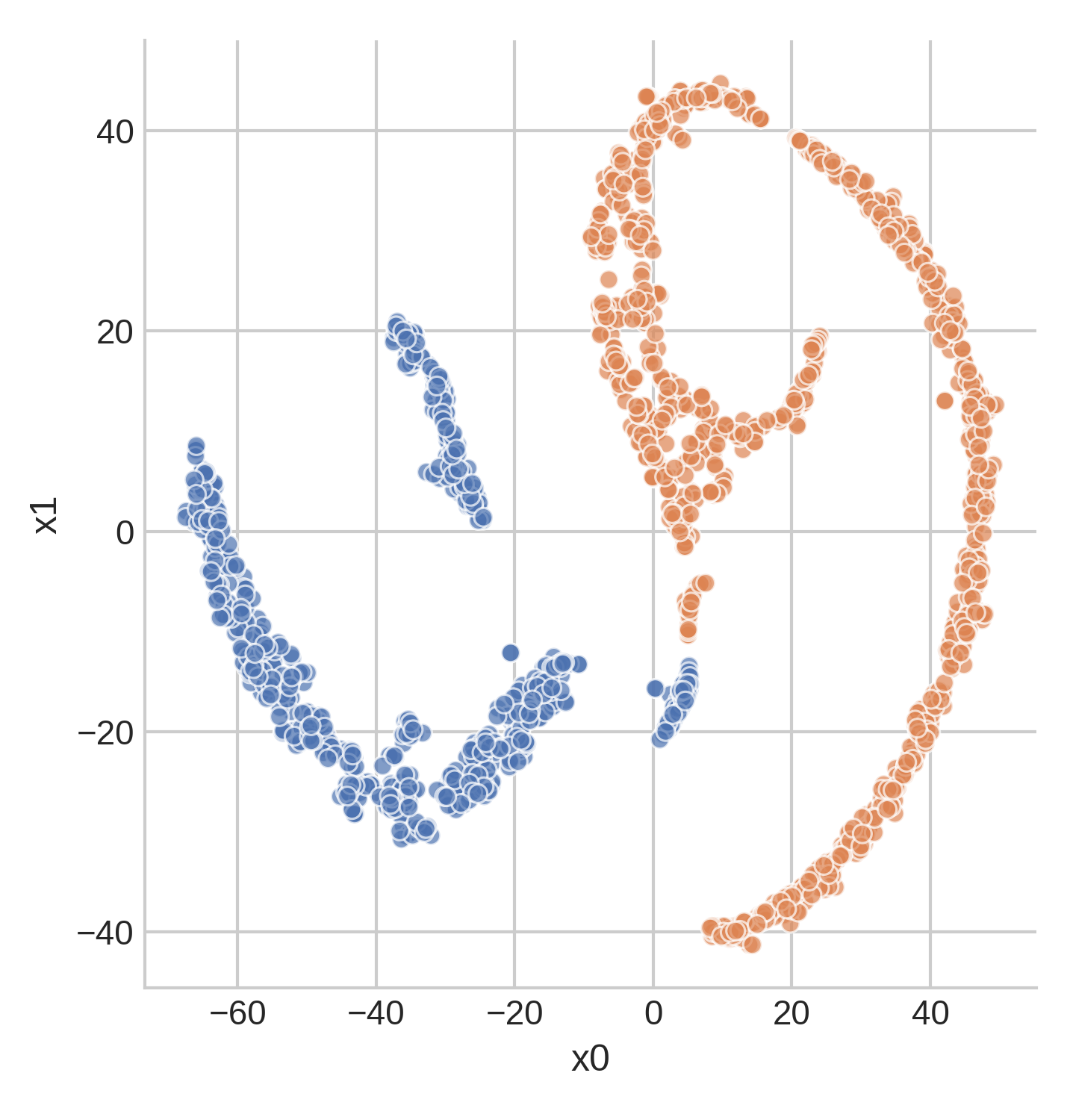}
    \caption{Model trained with a GRL ($\lambda=50$)}
    \label{fig:tsne-unbiased}
  \end{subfigure}
    ~
  \begin{subfigure}[t]{0.3\columnwidth}
    \includegraphics[width=\columnwidth]{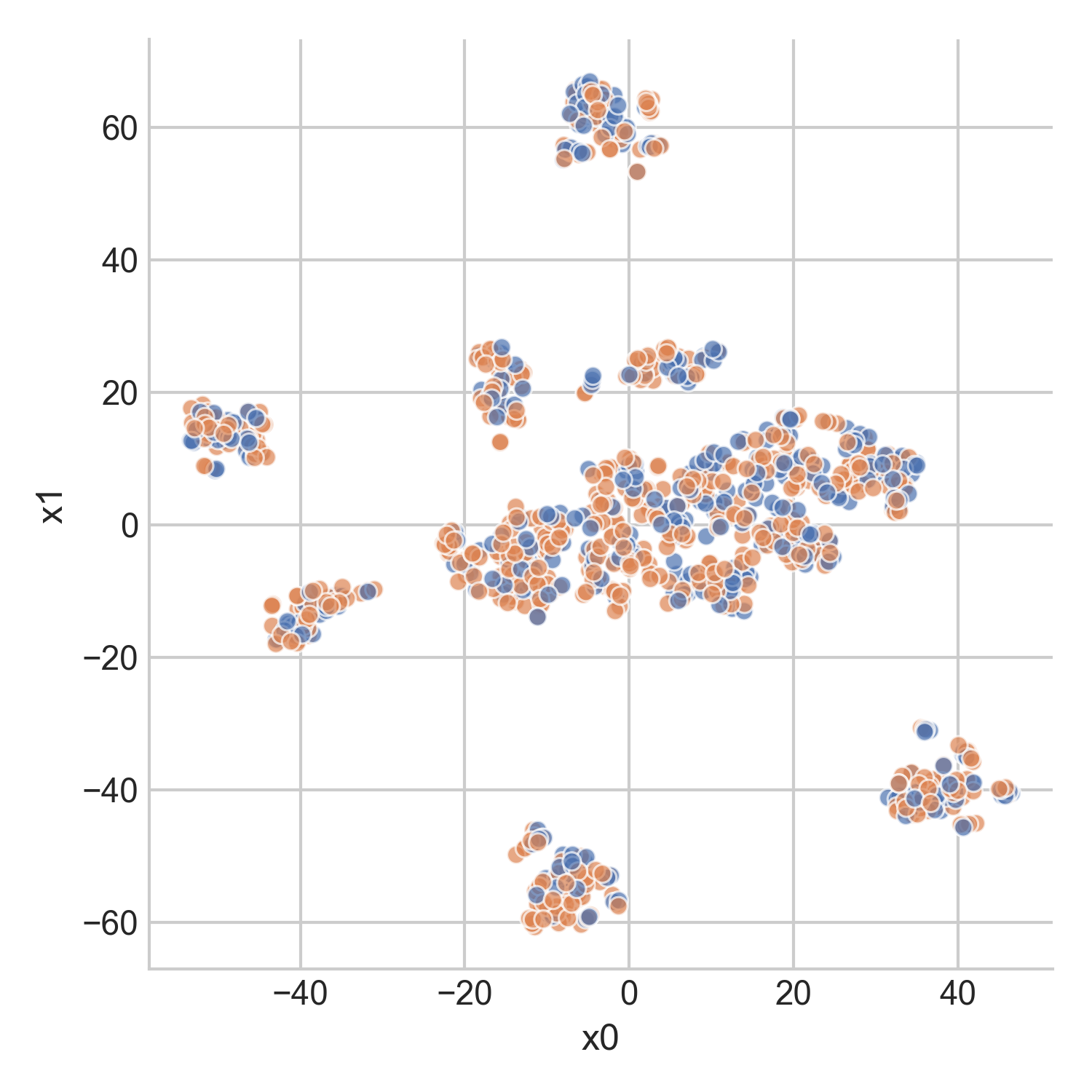}
    \caption{Model trained with our framework}
    \label{fig:tsne-unbiased-ours}
  \end{subfigure}
  \caption{T-SNE dimensionality reduction of the activations in the last hidden layer on the held-out COMPAS test set. Distinct colors are used for the reported race of individuals in the dataset: either African-American \statcirc{paper_blue} or Caucasian \statcirc{paper_orange}.}\label{fig:tsne}
\end{figure}

For each individual for the COMPAS test set, all three models derive a representation in the last hidden layer, on which
we applied a t-SNE dimensionality reduction for a two-dimensional visualisation.

The model without fairness constraints (\autoref{fig:tsne-biased}) has slight separation with regard to the protected attribute, 
but it is clearly separable in the representation from the model trained with a GRL (\autoref{fig:tsne-unbiased}).
This is also shown by retraining a one-layer perceptron on these representation. The model that was originally trained to predict only recidivism could be used to classify the protected attribute race with $AUC=0.71$. 
The adversarial branch $h_a$ that was trained simultaneously has an $AUC=0.44$
As we mentioned earlier, this branch can be limited in predicting the protected attribute $A$.
Which is the case here, as an independent perceptron has $AUC=0.92$. 

Here, we demonstrated that the hidden representation obtained by gradient reversal, not only still contains information about the protected attribute, but contains a stronger signal. Our architecture that joins `adversarial fairness', also called the Reader, and `adversarial learning', or the Feeder, (see~\autoref{fig:architecture}) leverages utility- {\em and} fairness-focused methods in a better way than the modification of the model alone. By injecting noise with the adversarial Feeder, our framework successfully mitigated this unfair representation, as shown in \autoref{fig:tsne-unbiased-ours}.

\subsection{Effect of adversarial fraction}\label{ss:adv-fraction}

\begin{figure}[tbh]
  \centering
  \centerfloat
  \begin{subfigure}[t]{0.34\columnwidth}
    \includegraphics[width=\columnwidth]{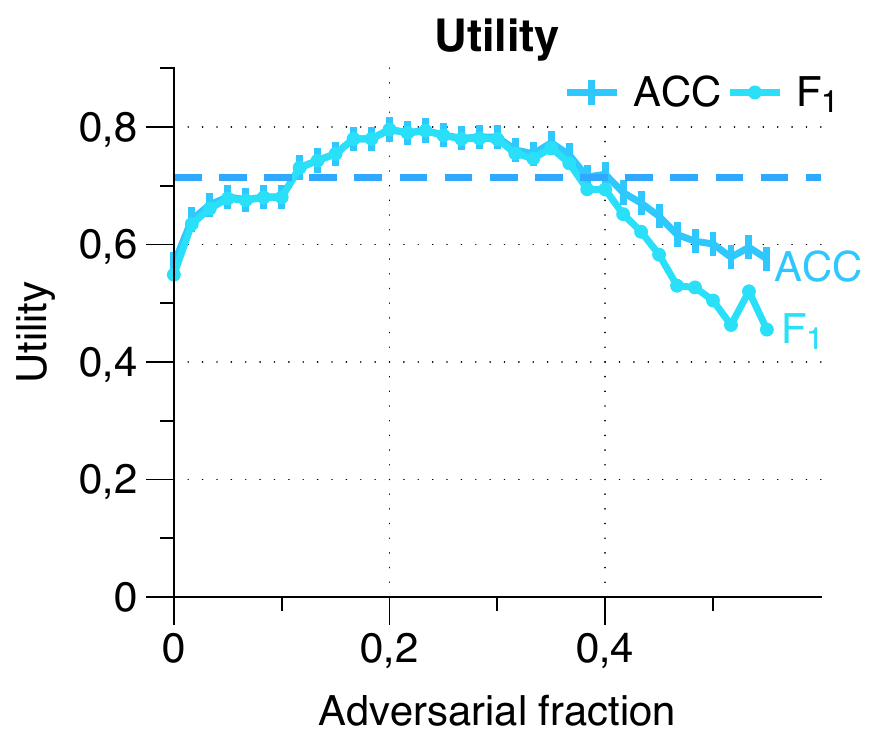}
    \subcaption{Utility}
  \end{subfigure}
  \begin{subfigure}[t]{0.34\columnwidth}
    \includegraphics[width=\columnwidth]{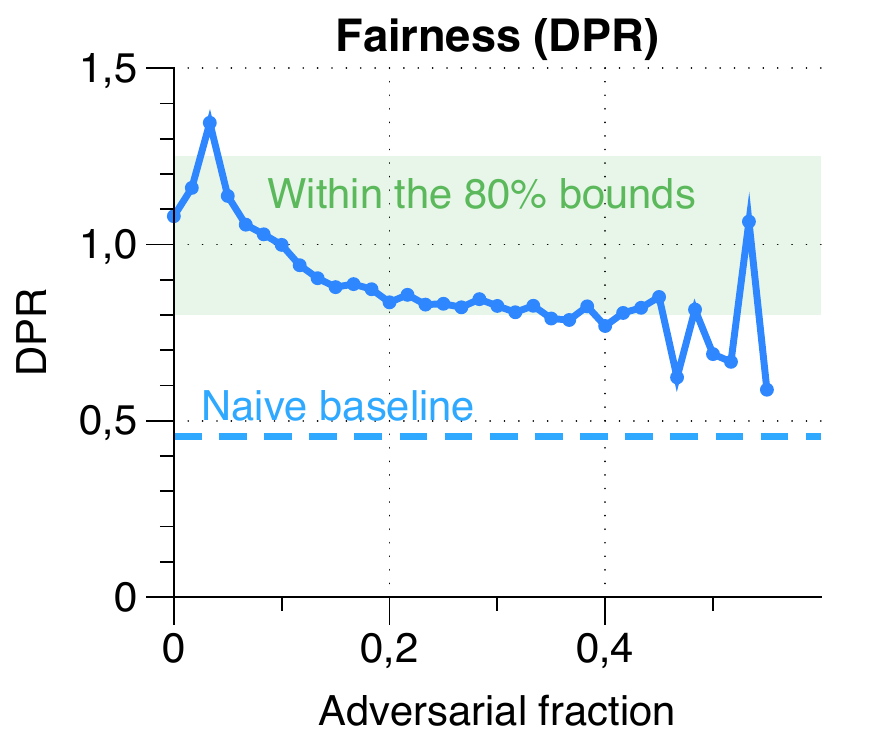}
    \subcaption{DPR}
  \end{subfigure}
  \begin{subfigure}[t]{0.34\columnwidth}
    \includegraphics[width=\columnwidth]{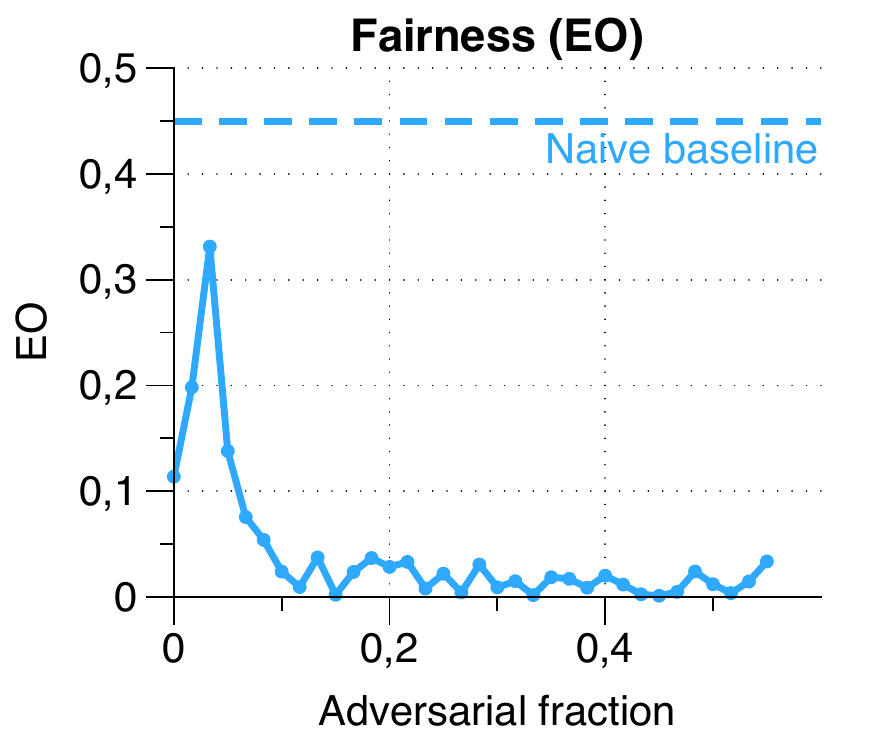}
    \subcaption{EO}
  \end{subfigure}
  \caption{Fairness and utility measures after each attack iteration on COMPAS (Batch size of 1024, $\lambda=100$, epochs=100, 50 adversarial points per iteration)}
  \label{fig:compas-history}
\end{figure}

\autoref{fig:compas-history} displays the effect of the adversarial fraction in the training dataset on COMPAS. 
When adversarial examples (equivalent to 25\% of the training set size) are added to the training set, the utility is maximal.
With higher fractions, the utility decreases and the development of the DP ratio fluctuates. This could stem from the minimax formulation, where a small fraction (i.e., 25\%) helps optimize better for this saddle point, but higher fractions only add noise. We use this fraction for all further experiments, in future work this could be automated with a custom stopping criterion.

\subsection{Benchmark results}\label{ss:results}

\begin{table}[ht!]
\centering
\caption{Results on the three datasets. An obelisk ($\dagger$) show results reported by original papers. Results of classifiers without fairness constraints are reported as a baseline. Best results are in bold typeface. An asterisk ($*$) indicates a division by zero.}\label{tab:results}
\resizebox{0.85\linewidth}{!}{
\begin{tabular}{@{}llllll@{}}
\toprule
Model                         & ACC                      & F1    & DP             & DPR 
& EO    \\ \midrule
\textbf{Adult} &                    &       &  &                   &       \\ 
Baseline without fairness constraints~~               & \textbf{0.839}$~\pm~$0.009          & \textbf{0.763} & 0.173          & 

0.296
& 0.096 \\ 
GRL             & 0.612$~\pm~$0.012          & 0.518 & 0.059          & 

1.931
& \textbf{0.061} \\
NBF (NB)~\citep{calders2010}  & 0.773$^\dagger$                    & ---      & \textbf{0.000}$^\dagger$ & ---                  & ---      \\
NBF (EM)~\citep{calders2010}  & 0.801$^\dagger$                    & ---   & 0.001$^\dagger$          & ---                  & ---      \\
Grad-Pred~\citep{raff2018}    & 0.754$^\dagger$                    & ---      & \textbf{0.000}$^\dagger$ & ---                  & ---      \\
FF~\citep{raff2018fair}       & 0.753$^\dagger$                    & ---      & \textbf{0.000}$^\dagger$ & ---                  & ---      \\
LFR~\citep{zemel2013learning} & 0.702$^\dagger$                    & ---      & 0.001$^\dagger$          & ---                  & ---      \\
Ours                          & 0.814$~\pm~$0.009~~ & 0.689 & 0.031          & 

\textbf{0.784}
& 0.179 \\ \midrule
\textbf{German Credit} &                    &       &  &                   &       \\ 
Baseline without fairness constraints~~              & 0.705 $~\pm~$ 0.063    & 0.624          & 0.018          & 

0.929
& 0.198 \\ 
GRL                                                       & 0.710  $~\pm~$ 0.063            &  0.415        & \textbf{0.000}          & $*$    & \textbf{0.000} \\
Grad-Pred~\citep{raff2018}    & 0.675$^\dagger$                    & ---      & 0.001$^\dagger$ & ---                  & ---      \\
FF~\citep{raff2018fair}       & 0.700$^\dagger$                    & ---      & \textbf{0.000}$^\dagger$ & ---                  & ---      \\
LFR~\citep{zemel2013learning} & 0.591$^\dagger$                    & ---      & 0.004$^\dagger$          & ---                  & ---      \\
Ours                                                                    & \textbf{0.730}$~\pm~$0.062    & \textbf{0.640} & 0.006 & 

\textbf{0.971}
& 0.175 \\ \midrule
\textbf{COMPAS} &                    &       &  &                   &       \\ 
Baseline without fairness constraints~~              & 0.715             & 0.709          & 0.466          & 

2.192
& 0.449 \\ 
GRL                                                       & 0.567             & 0.549          & 0.057          & 

\textbf{0.926}
& 0.114 \\
COMPAS risk predictions~\citep{angwinMachineBiasThere2016}                                                                 & 0.655$~\pm~$0.029 & 0.654          & 0.289          & 1.829         & \textbf{0.000} \\
Preference-based fairness~\citep{zafarParityPreferencebasedNotions2017} & 0.675$^\dagger$             &  ---              & 0.380$^\dagger$          & ---         &  ---     \\
Ours                                                                    & \textbf{0.794}    & \textbf{0.793} & \textbf{0.026} & 

0.840
& 0.008 \\
\bottomrule
\end{tabular}
}
\end{table}

\autoref{tab:results} presents our results on the three datasets.
We compare them with (i) a baseline without fairness goals, \emph{i.e.}, a neural network without any particular control on fairness aspects, (ii) a re-implementation of the GRL~\citep{raff2018, adel2019, ganin2016domain} and (iii) the reported results from other works that incorporate fairness and cover a wide range of learning algorithms: Naive Bayes~\citep{calders2010}, random forests~\citep{raff2018fair}, SVMs~\citep{zafarParityPreferencebasedNotions2017} and neural networks~\citep{raff2018, zemel2013learning}.
The models' utility was evaluated by binary classification accuracy and macro-averaged $F_1$ score; the latter highlights some issues when dealing with class imbalances, as is the case for German Credit.

Fairness is evaluated with demographic parity, both as an absolute difference (DP) and as a ratio (DPR), and equal opportunity (EO).
\citet{adel2019} also report results on both COMPAS and Adult but use a different setup for the Adult dataset. 
For COMPAS, the reported results (as well as their unfair baseline) are significantly higher than in our experiments, which we could replicate only when classifying high-risk individuals. 
To make a meaningful comparison, we also include our replication of \emph{FAD}~\citep{adel2019} as \emph{GRL}.

The utility of our framework is the highest on the German Credit and COMPAS datasets, even surpassing the baseline model. On Adult, we achieve the highest utility of any model with fairness constraints. These results show that our model has only a very limited impact on the utility of the classifier, and it can even contribute to the training as shown in \autoref{fig:compas-history}. Note that on German Credit, a majority classifier would achieve 70\% accuracy already, hence the inclusion of the  $F_1$~score.

Regarding fairness evaluation, our framework gives the best results for COMPAS when considering DP. 
It also increases fairness as measured by DPR, which is the only one of the considered measures that indicates the ``direction'' of unfairness. More fairness is sometimes given by an {\em increase} towards parity (DPR=1) for the disadvantaged group: for the German Credit dataset, their chances of getting a loan increase. 
In COMPAS,
the baseline has a EO of 2.192, the ``bias against blacks'' \citep{angwinMachineBiasThere2016} {\em decreases} substantially with our model.
For GRL, the near-equality of DPR (0.926) appears fairer, but this is not the case for DP and EO, where we observe an EO of 0.449 for GRL versus 0.008 for our model. 

\section{Code}
We release an open source implementation--- under the MIT licence---of our framework at \url{https://github.com/iPieter/ethical-adversaries}.

\section{Summary, conclusions and future work}\label{sec:conclusion}
\input{conclusion}

\section*{Acknowledgements}

Pieter Delobelle was supported by the Research Foundation - Flanders under EOS No. 30992574 and received funding from the Flemish Government under the “Onderzoeksprogramma Artificiële Intelligentie (AI) Vlaanderen” programme.
Gilles Perrouin is an FNRS Research Associate.
This research was partly supported by EOS Verilearn project grant no. O05518F-RG03.
We also want to thank the secML developers from the PRALab (Pattern Recognition and Applications Laboratory, University of Cagliari, Sardegna, Italy) for having answered our numerous questions and helping us in using their newly developed library.

\bibliographystyle{splncsnat}
\bibliography{paper, more-refs}

\end{document}

%% file: absrtact.tex
Machine learning is being integrated into a growing number of critical systems with far-reaching impacts on society. 
Unexpected behaviour and unfair decision processes are coming under increasing scrutiny due to this widespread use and its theoretical considerations. 
Individuals, as well as organisations, notice, test, and criticize unfair results to hold model designers and deployers accountable. 
We offer a framework that assists these groups in mitigating unfair representations stemming from the training datasets.
Our framework relies on two inter-operating adversaries to improve fairness.
First, a model is trained with the goal of preventing the guessing of protected attributes' values while limiting utility losses. This first step optimizes the model’s parameters for fairness.
Second, the framework leverages evasion attacks from adversarial machine learning to generate new examples that will be misclassified. 
These new examples are then used to retrain and improve the model in the first step.
These two steps are iteratively applied until a significant improvement in fairness is obtained. 
We evaluated our framework on well-studied datasets in the fairness literature — including COMPAS — where it can surpass other approaches concerning demographic parity, equality of opportunity and also the model’s utility. 
We also illustrate our findings on the subtle difficulties when mitigating unfairness and highlight how our framework can assist model designers.

%% file: architecture.tex
Our main contribution is a framework that joins evasion attacks (see \autoref{sec:adv-ml}) and fair neural networks (see \autoref{sec:fair-NN}) to improve the overall fairness of the system. Thus, it relies on two types of ethical adversaries: (i) a \textit{Feeder} that uses evasion attacks to create examples highlighting unfair representation of a certain population and (ii) an \textit{adversarial Reader} that is trying to predict the protected attributes of interest (age, gender, race, etc.). In addition of exhibiting fairness issues in the data and in the trained model, our framework leverages gradient reversal to minimise the ability of the reader to guess protected attributes ultimately yielding a fairer ML model without sacrificing utility.

\begin{figure}[tbh]
  \centering
  \includegraphics[width=0.9\textwidth]{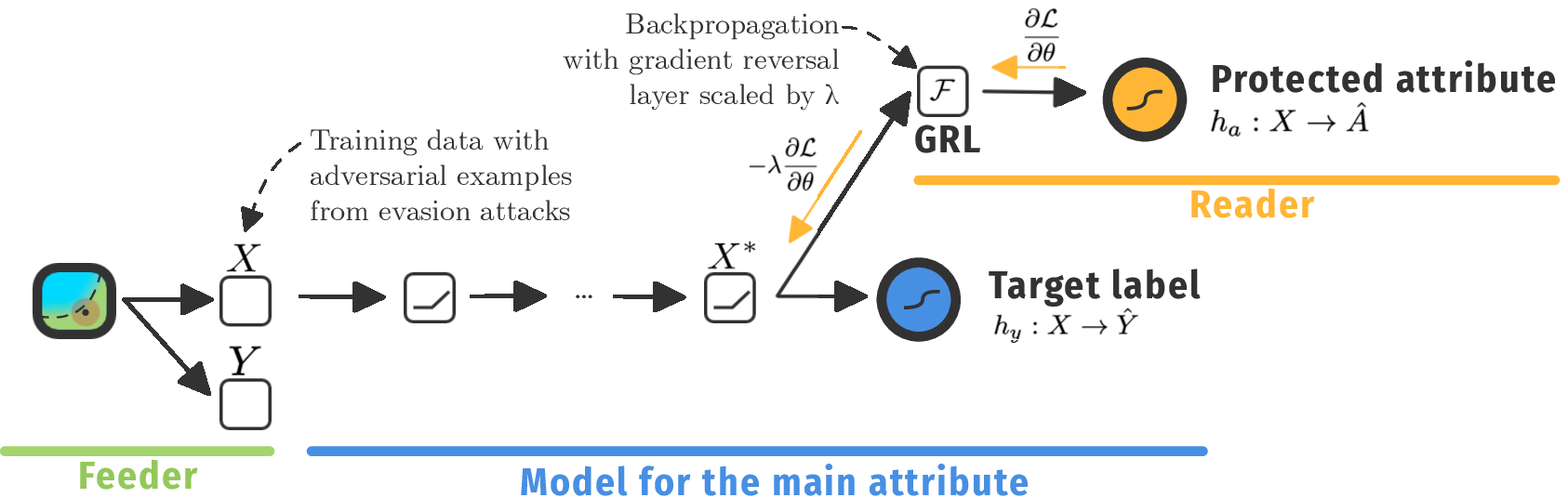}
  \caption{Ethical adversaries architecture: adversarial feeder on the left, and integrated adversarial reader on the right.}
  \label{fig:architecture} 
\end{figure}

\autoref{fig:architecture} presents the global architecture. Our network follows a typical architecture with a GRL (discussed  in  \autoref{sec:fair-NN} and  is  represented  by the  Reader). 
The Feeder, on the left part, performs evasion attacks as discussed in \autoref{sec:adv-ml}. 
Both adversaries interact with each other in an iterative manner—which is the main difference between our framework and GANs \citep{goodfellowGenerativeAdversarialNets2014a}. 
To achieve better fairness and utility outcomes, the process, that consists of two steps, can be performed multiple times.

The first step starts with a trained neural network (target label in \autoref{fig:architecture}) predicting a main attribute $Y$. In this network, the adversarial Reader adds a second branch that tries to predict a protected attribute $A$ while the gradient reversal layer strives to minimise the confidence of the Reader to predict $A$. Additionally, as we discussed in \autoref{sec:fair-NN}, during the backward pass, a hyperparameter~$\lambda$ contributes to prioritize the utility versus the adversarial branch of the network. The model is trained with the joint loss of the original prediction target and the protected attribute. 

In a second step, the Feeder, on the left part, performs evasion attacks as discussed in \autoref{sec:adv-ml}. 
The Feeder creates a set of adversarial points from an approximation of the target model, a.k.a a surrogate model, that is constructed on the same dataset as the model under attack. 
Our surrogate model is an SVM which it uses a radial basis (RBF) kernel function to cope with different level of model complexity.
We selected this kernel since preliminary results on COMPAS showed that it is expressive enough and \citet{biggio2013evasion} detailed how evasion attacks can be directly applied to SVMs with RBF kernels.
The Feeder performs multiple evasion attacks on the surrogate function to generate adversarial examples that are similar to the training examples, but are wrongly classified.

For each iteration of this two-step process, adversarial examples are generated and included in the training set for adversarial retraining.
Each adversarial example is added to the training set with the same label as the original example from which it was generated.
The effect of the ratio of adversarial points in the dataset—the adversarial fraction—is further analyzed empirically in \autoref{ss:adv-fraction}.

In terms of performance, constructing a surrogate classifier is the limiting factor. Using SVMs implies that the time complexity of the entire framework is $\mathcal{O}(n^3)$ with $n$ the number of data points. The impact of adversarial attacks is linear on the overall complexity. 
But note that adversarial retraining may drastically increase time to compute a separating function since included adversarial examples make the separation more difficult to find, or on the contrary, may not affect the function at all, if few adversarial examples are included.

Both reading and feeding steps are run successively until we achieve better fairness and utility outcomes, which we demonstrate in \autoref{ss:results}. A key benefit of this process is that we prevent the Reader from learning biased representations, since these features cannot be used as proxies for the protected attribute anymore.

%% file: conclusion.tex
In this paper, we presented a novel architecture for integrating fairness constraints in machine learning models. Our architecture consists of two adversaries: (i) an adversarial reader that evaluates fairness constraints during model training and attempts to enforce them, and (ii) an adversarial feeder that performs iterative evasion attacks to discover previously uncovered regions in the input space. 
We evaluated our architecture on three well-studied datasets and showed that it can deliver high utility to models while satisfying fairness constraints. On COMPAS, we illustrated that our architecture yields a model that surpasses an unfair baseline regarding the utility (accuracy and $F_1$ score) and fairness. We provide evidence that gradient reversal alone is not sufficient (it might even be detrimental) but that our combination of adversaries leads to intrinsically fairer models.  

There is room for future work. First, we may optimize the runtime execution of the technique via faster learning of surrogate models. Second, we could use the target model directly instead of a surrogate classifier to support adversarial attacks and assess if transferability properties hold for fairness constraints. This requires heavyweight modifications of the secML framework to allow multiple output values in neural networks. Third, one could define constraints involving multiple features.
Enforcing these \textit{domain-specific} constraints during attack generation raises questions on the representation of the feature space and optimal convergence of the algorithms.
Fourth, our framework is evaluated against allocational harms. More subtle differences— like a difference in the model's performance—are also affecting social groups.
With some minor modifications, we suspect that these types of unfairness can be addressed with our framework. 
Finally, we would like to generate the most dissimilar examples possible to ensure good coverage of the unseen feature space with a minimal number of attacks.